\pdfoutput=1
\documentclass[11pt,a4paper]{article}
\usepackage{emnlp2021}
\usepackage[american]{babel}
\usepackage[T1]{fontenc}
\usepackage{times}
\usepackage{url}
\usepackage{microtype}
\usepackage{lipsum}
\usepackage[utf8]{inputenc}
\usepackage{soul}

\usepackage{booktabs}

\usepackage[hang,flushmargin]{footmisc}
\newcommand{\Ni}{(1)~}
\newcommand{\Nii}{(2)~}
\newcommand{\Niii}{(3)~}
\newcommand{\Niv}{(4)~}
\newcommand{\Nv}{(5)~}

\usepackage{graphicx}
\DeclareGraphicsExtensions{.pdf,.ai,.jpg,.png}
\setkeys{Gin}{pagebox=artbox}% Use artbox instead of mediabox default. Options: mediabox, cropbox, bleedbox, trimbox, artBox. (shell: pdfinfo -box <pdf-file>)

\graphicspath{{.}}

\newcommand{\hwfigure}[3][scale=1.0]{%
    \begin{figure*}[tb]
        \centering
        \includegraphics[#1]{#2}
        \vglue -1.0ex plus 0.0ex minus 0.5ex
        \caption{#3}\label{#2}%
    \end{figure*}
}

\newsavebox\bscombox
\newcommand{\bscom}[3][]{%
  \sbox{\bscombox}{\fontsize{8}{9}\selectfont#1#2#3}
  \noindent
  \st{#2}{\fontsize{8}{9}\selectfont
    \color{blue}#3\ifx\\#1\\\else{\fontsize{8}{9}\selectfont\color{violet}[#1]}\fi
    }
  }

\newcommand\blfootnote[1]{%
  \begingroup
  \renewcommand\thefootnote{}\footnote{#1}%
  \addtocounter{footnote}{-1}%
  \endgroup
}

\begin{document}
\title{{\textsc{Summary Explorer}}\\ Visualizing the State of the Art in Text Summarization}

\newcommand{\lei}{\textsuperscript{$\dagger$}}
\newcommand{\den}{\textsuperscript{$\ddagger$}}

\author{%
Shahbaz Syed \lei*
\qquad Tariq Yousef \lei*
\qquad Khalid Al-Khatib \lei \\[1.5ex]
\bfseries Stefan Jänicke \den \hspace{1.5ex}
\bfseries Martin Potthast \lei \\[1.5ex]
\hspace{-5pt}\lei{}Leipzig University\quad\den{}University of Southern Denmark\\
{\small\tt{<shahbaz.syed@uni-leipzig.de>
\small\tt<tariq.yousef@uni-leipzig.de>
}}}
\date{}

\maketitle
\begin{abstract}
This paper introduces {\small\textsc{Summary Explorer}}, a new tool to support the manual inspection of text summarization systems by compiling the outputs of 55~state-of-the-art single document summarization approaches on three benchmark datasets, and visually exploring them during a qualitative assessment.
The underlying design of the tool considers three well-known summary quality criteria (coverage, faithfulness, and position bias), encapsulated in a guided assessment based on tailored visualizations. The tool complements existing approaches for locally debugging summarization models and improves upon them. The tool is available at {\url{https://tldr.webis.de/}}.

\blfootnote{* Equal contribution.}
\end{abstract}

\section{Introduction}

Automatic text summarization is the task of generating a summary of a long text by condensing it to its most important parts. This longstanding task originated in automatically creating abstracts for scientific documents \cite{luhn:1958}, and later extended to documents such as web pages \cite{salton:1994} and news articles \cite{wasson:1998}.

There are two paradigms of automatic summarization: \emph{extractive} and \emph{abstractive}. The former extracts important information from the to-be-summarized text, while the latter additionally involves paraphrasing, sentence-fusion, and natural language generation to create fluent summaries. 
Neural summarization approaches trained on large-scale datasets have significantly advanced both paradigms by improving the overall document understanding and text generation capabilities of the models to generate fluent summaries.

\enlargethispage{\baselineskip}
Currently, the progress in text summarization is tracked primarily using {\em automatic evaluation} with ROUGE \cite{lin:2004} as the de facto standard for quantitative evaluation. ROUGE has proven effective for evaluating extractive systems, measuring the overlap of word n-grams between a generated summary and a reference summary (ground truth). Still, it only provides an approximation of a model's capability to generate summaries that are lexically similar to the ground truth. Moreover, ROUGE is unsuitable for evaluating abstractive summarization systems, mainly due to its inadequacy in capturing all semantically equivalent variants of the reference \cite{ng:2015,kryscinski:2019,fabbri:2021}.
Besides, a reliable automatic evaluation of a summary is challenging \cite{lloret:2018} and strongly dependent on its purpose\cite{jones:1999}. 

A robust method to analyze the effectiveness of summarization models is to manually inspect their outputs from individual perspectives such as coverage of key concepts and linguistic quality. However, manual inspection requires obtaining the outputs of certain models, delineating a guideline that comprises particular assessment criteria, and ideally utilizing proper visualization techniques to examine the outputs efficiently.

\hwfigure[width=\textwidth]{summary-explorer-overview}{Overview of {\small\textsc{Summary Explorer}}. Its guided assessment process works in four steps:
\Ni
corpus selection,
\Nii
quality aspect selection,
\Niii
model selection, and
\Niv
quality aspect assessment.
Exemplified is the assessment of the content coverage of the summaries of four models for a source document from the CNN/DM corpus. For each summary sentence, its two most related source document sentences are highlighted on demand.}

To this end, we present {\small\textsc{Summary Explorer}} (Figure~\ref{summary-explorer-overview}), an online interactive visualization tool that assists humans (researchers, experts, and crowds) to inspect the outputs of text summarization models in a guided fashion. Specifically, we compile and host the outputs of several state-of-the-art models (currently~55) dedicated to English single-document summarization. These outputs cover three benchmark summarization datasets comprising semi-extractive to highly abstractive ground truth summaries. The tool facilitates a {\em guided} visual analysis of three important summary quality criteria: \emph{coverage}, \emph{faithfulness}, and \emph{position bias}, where tailored visualizations for each criterion streamline both absolute and relative manual evaluation of summaries. Overall, our use cases (see Section~\ref{analysis}) demonstrate the ability of {\small\textsc{Summary Explorer}} to provide a comparative exploration of the state-of-the-art text summarization models, and to discover interesting cases that cannot likely be captured by automatic evaluation.
\section{Related Work}

Leaderboards such as Paperswithcode,%
\footnote{\url{https://paperswithcode.com/task/text-summarization}}
Explaina\-Board%
\footnote{\url{http://explainaboard.nlpedia.ai/leaderboard/task-summ/}}
and NLPProgress%
\footnote{\url{https://nlpprogress.com/english/summarization.html}}
provide an overview of state of the art in text summarization mainly according to ROUGE. These leaderboards simply aggregate the scores as reported by the models' developers, where the reported scores can be obtained using different implementations. Hence, a fair comparison become less feasible. For instance, the Bottom-Up model \cite{gehrmann:2018} uses a different implementation of ROUGE,%
\footnote{\url{https://github.com/sebastianGehrmann/rouge-baselines}}
compared to the BanditSum model \cite{dong:2018}.%
\footnote{\url{https://github.com/pltrdy/rouge}}
Besides, for a qualitative comparison of the models, one needs to manually inspect the generated summaries, which are missing from such leaderboards.

To address these shortcomings, VisSeq~\cite{wang:2019} aids developers to locally compare their model's outputs with the ground truth, providing lexical and semantic comparisons along with statistics such as most frequent n-grams and sentence score distributions. LIT~\cite{tenney:2020} provides similar functionality for a broader range of NLP tasks, implementing a work-bench-style debugging of model behavior, including visualization of model attention, confusion matrices, and probability distributions. Closely related to our work is SummVis~\cite{vig:2021}, the recently published tool that provides a visual text comparison of summaries with a reference summary as well as a source document, facilitating local debugging of hallucinations in the summaries.

{\small\textsc{Summary Explorer}} draws from these developments and adds three missing features:
\Ni
Quality-criteria-driven design. Based on a careful literature review of qualitative evaluation of summaries, we derive three key quality criteria and encode them explicitly in the interface of our tool. Other existing tools render these criteria implicit in their underlying design.
\Nii
A step-by-step process for guided analysis.  From the chosen quality criteria, we formulate concise and specific questions needed for a qualitative evaluation, and provide a tailored visualization for each question. While previous tools utilize visualization and enable users to (de)activate certain features, they oblige the users to figure out the process themselves, which can be overwhelming to non-experts.
\Niii
Compilation of the state of the art. We collect the outputs of more than 50~models on three benchmark datasets providing a comprehensive overview of the progress in text summarization.

{\small\textsc{Summary Explorer}}  complements these tools and also provides direct access to the state of the art in text summarization, encouraging rigorous analysis to support the development of novel models.
\section{\kern-0.25em Designing Visual Summary Exploration}
\label{summary-explorer}

The design of {\small\textsc{Summary Explorer}} derives from first principles, namely the three quality criteria \emph{coverage}, \emph{faithfulness}, and \emph{position bias} of a summary in relation to its source document. These high-level criteria are frequently manually assessed throughout the literature. Since their definitions vary, however, we derive from each criterion a total of six specific aspects that are more straightforwardly operationalized in a visual exploration (see Figure~\ref{summary-explorer-overview}, Step~2). To render the aspects more directly accessible to users, each is ``clarified'' by a guiding question that can be answered by a tailored visualization. Below, the three quality criteria are discussed, followed by the visual design.

\subsection{Summary Quality Criteria}

\paragraph{Coverage}
A primary goal of a summary is to capture the important information from its source document. Accordingly, a standard practice in summary evaluation is to assess its coverage of the key content \cite{paice:1990,mani:2001,jones:1999}. In many cases, a comparison to the ground truth (reference) summary can be seen as a proxy for coverage, which is essentially the core idea of ROUGE. However, since it is hard to establish an ideal reference summary \cite{mani:1999}, a comparison against the source document is more meaningful. Although an automatic comparison against it is feasible \cite{nenkova:2013,shafieibavani:2018}, deciding what is \emph{important} content is highly subjective \cite{peyrard:2019}. Therefore, authors resort to a manual comparison instead \cite{hardy:2019}.
We operationalize coverage assessment by visualizing a document's overlap in terms of content, entities, and entity relations with its summary. Content coverage refers to whether a summary condenses information from all important parts of a document, measured by common similarity measures; entity coverage contrasts the sets of named entities identified in both summary and document; and relation coverage does the same, but for extracted entity relations.

\paragraph{Faithfulness}
A more recent criterion that gained prominence especially in relation to neural summarization is the faithfulness of a summary to its source document \cite{cao:2018,maynez:2020}. Whereas coverage asks if the document is sufficiently reflected in the summary, faithfulness asks the reverse, namely if the summary adds something new, questioning its appropriateness. Due to their autoregressive nature, neural summarization models have the unique property to ``hallucinate'' new content \cite{kryscinski:2020,zhao:2020}. This is what enables abstractive summarization, but also bears the risk of generating content in a summary that is unrelated to the source document. The only acceptable hallucinated content in a summary must be textually entailed by its source document, which renders an automatic assessment challenging \cite{falke:2019,durmus:2020}.
We operationalize faithfulness assessment by visualizing previously unseen words in a summary in context, aligned with the best-matching sentences of its source document.

\paragraph{Position bias}
Data-driven approaches, such as neural summarization models, can be biased by the domain of their training data and learn to exploit common patterns. For example, news articles are typically structured according to an ``inverted pyramid,'' where the most important information is given in the first few sentences \cite{purdue:2019}, and which models learn to exploit \cite{wasson:1998,kedzie:2018}. Non-news texts, such as social media posts, however, do not adopt this structure and thus require an unbiased consideration to obtain proper summaries \cite{syed:2019}.
We operationalize position bias assessment by visualizing the parts of a document that are the source of its summary's sentences, as well as the ones that are common among a set of summaries.

\subsection{Visual Design}

\begin{figure*}[!ht]
{\footnotesize (a)}\\
\includegraphics[width=\textwidth]{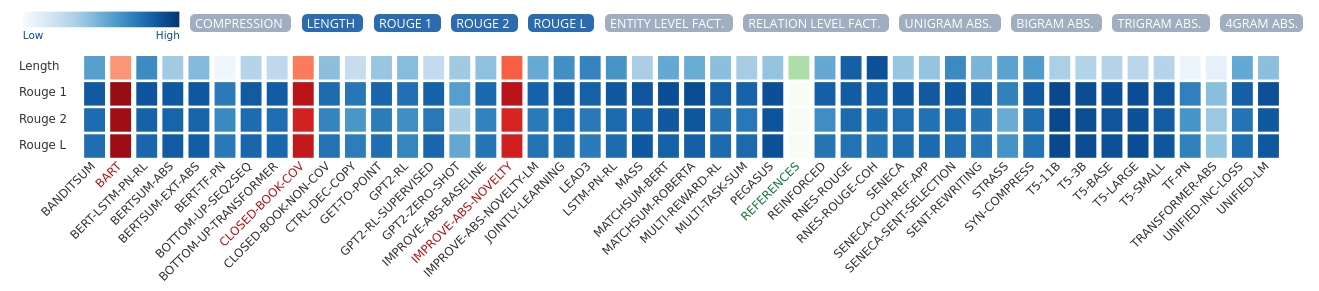}\\[-3ex]
% {\footnotesize
% (b)\hspace{15em}%
% (c)\hspace{1.5em}%
% (d)\hspace{1.5em}%
% (e)\hspace{2em}%
% (f)\par}
{\footnotesize (b)}\hspace{28em}{\footnotesize (g)}\\
\includegraphics[scale=0.25]{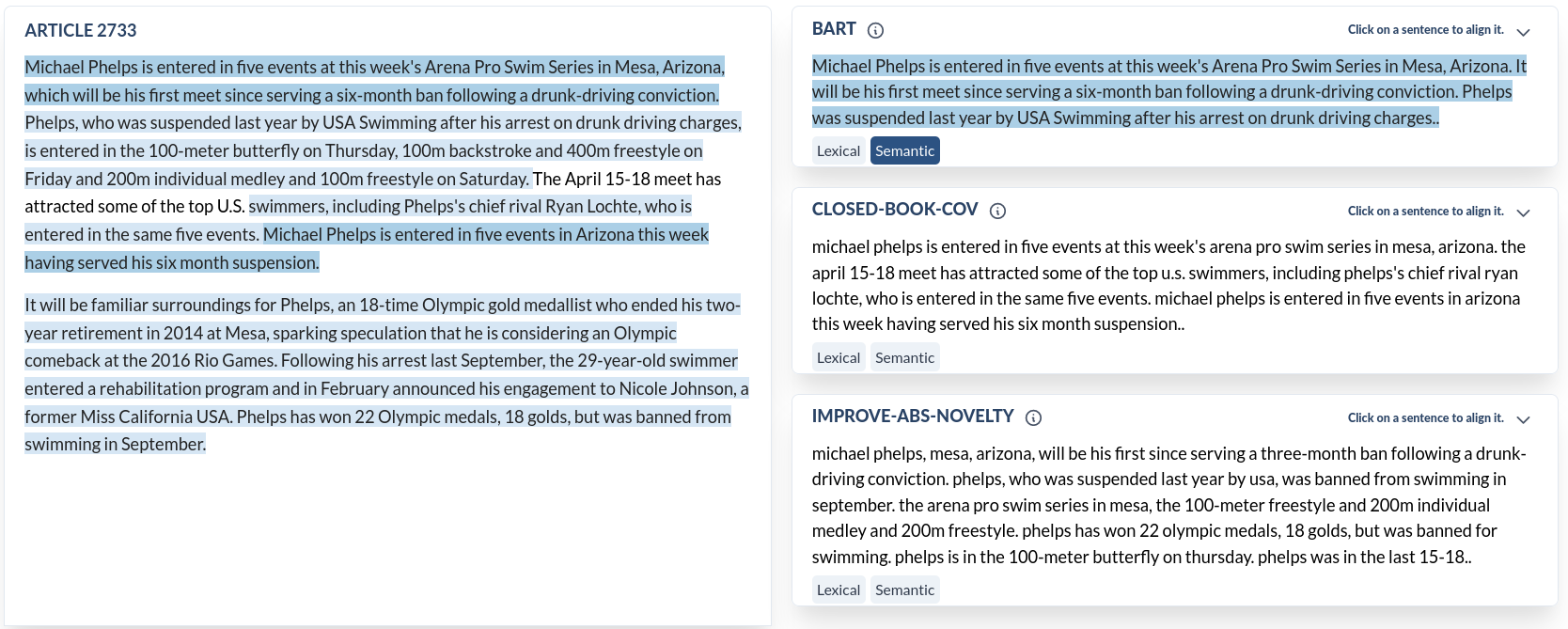}\\[-0.5ex]
{\footnotesize (c)}\\
\includegraphics[scale=0.25]{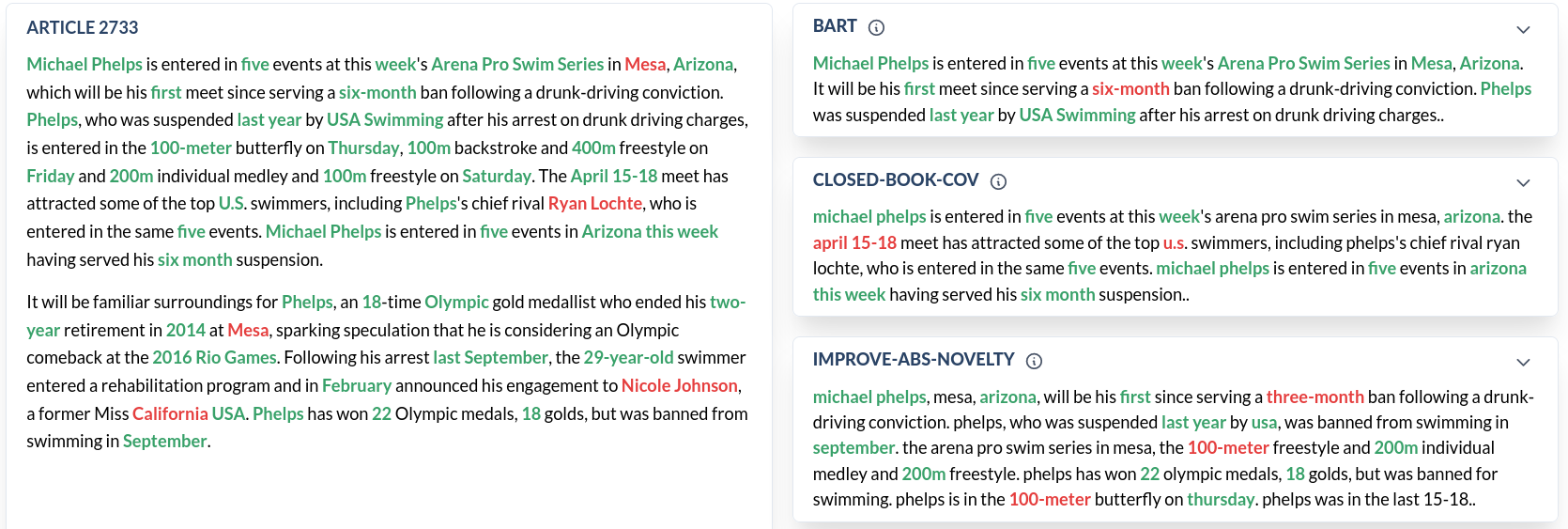}\\[-0.75ex]
{\footnotesize (d)}\hspace{12.8em}{\footnotesize (e)}\\
\includegraphics[scale=0.18]{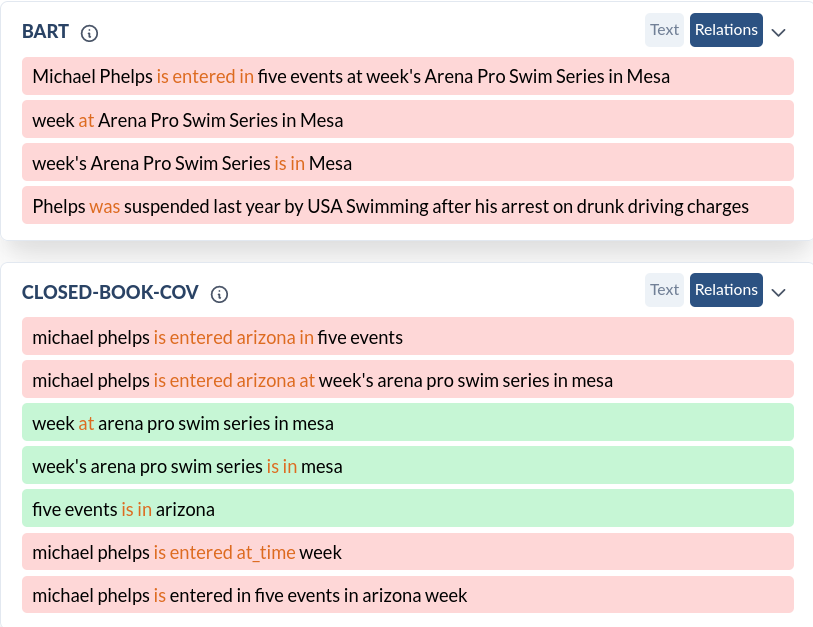}%
\includegraphics[scale=0.21]{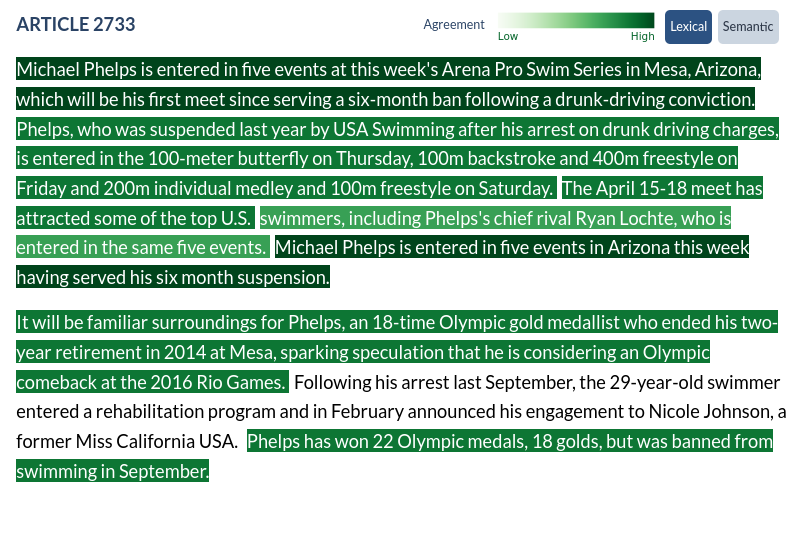}\\[-0.5ex]
{\footnotesize (f)}\\
\includegraphics[scale=0.18]{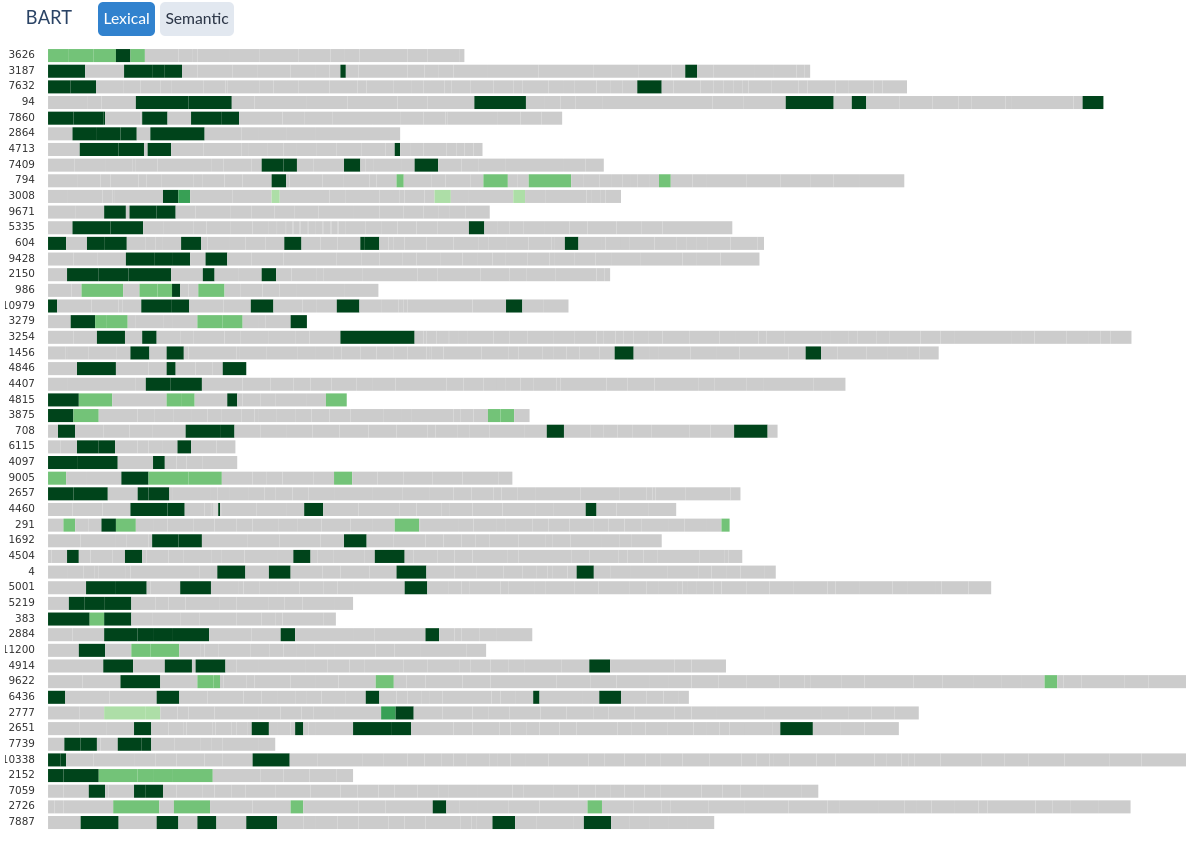}
\includegraphics[scale=0.18]{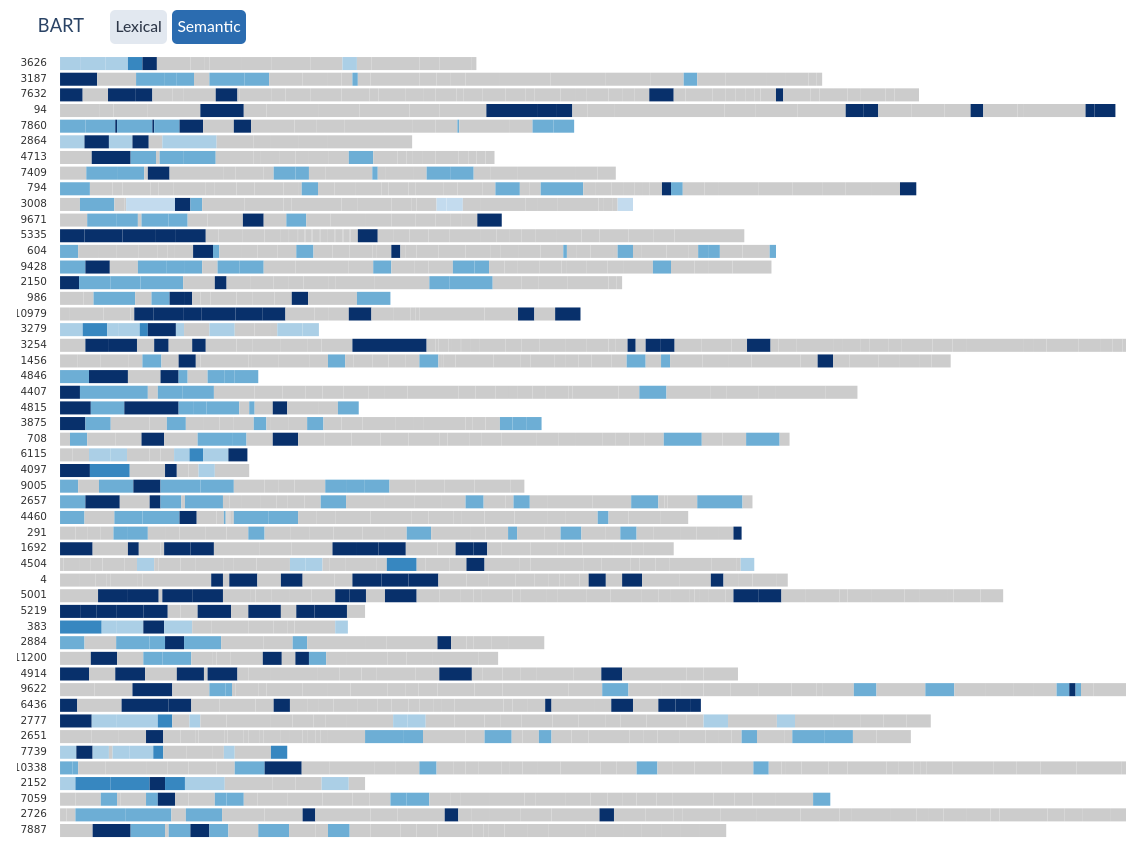}%
\raisebox{6ex}[0em][0em]{\includegraphics[scale=0.521]{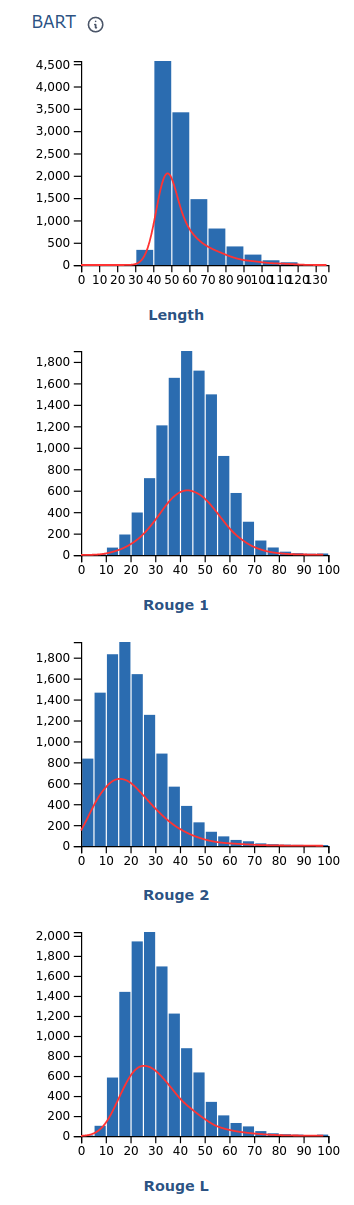}}
\caption{
(a)~Heatmap overview of 45~models for the CNN/DM corpus; ones selected for analysis are highlighted red.
Views for
(b)~the content coverage,
(c)~the entity coverage,
(d)~the relation coverage,
(e)~the position bias across models for a single document,
(f)~the position bias of a model across all documents as per lexical and semantic alignment,
(g)~the distribution of quantitative metric scores for a model.}
\label{summary-explorer-views}
\end{figure*}
\paragraph{Guided Assessment}
{\small\textsc{Summary Explorer}} implements a streamlined process to guide summary quality assessment, consisting of four steps (see Figure~\ref{summary-explorer-overview}).
\Ni
A benchmark dataset is selected.
\Nii
A list of available summary quality aspects is offered each with a preview of its tailored visualization and its interactive use.
\Niii
Applying Shneiderman's~(\citeyear{shneiderman:1996}) well-known Visual Information-seeking Mantra (``overview first, zoom and filter, then details-on-demand''), an overview of all models as a heatmap over averages of several quantitative metrics  is shown (Figure~\ref{summary-explorer-views}a), which enables a targeted filtering of the models based on their quantitative performance. The heatmap of average values paints only a rough picture; upon model selection, histograms of each model's score distribution for each metric are available.
\Niv
After models have been selected, the user is forwarded to the corresponding quality aspect's view.

\medskip
The visualizations for the individual aspects of the three quality criteria share the property that two texts need to be visually aligned with one another.%
\footnote{A visualization paradigm recently surveyed by \citet{yousef:2021}.}
Despite this commonality, we abstain from creating a single-view visualization ``stuffed'' with alternative options. We rather adopt a minimalistic design for the assessment of individual quality aspects.

\paragraph{Coverage View \normalfont (Figure~\ref{summary-explorer-views}b,c,d)}
Content coverage is visualized as alignment of summary sentences and document sentences at the semantic and lexical level in a full-text side-by-side view. Colorization indicates different types of alignments. For entity coverage (relation coverage), a corresponding side-by-side view lists named entities (relations) in a summary and aligns them with named entities (relations) in its source document. For unaligned relations, corresponding document sentences can be retrieved.

\paragraph{Faithfulness View \normalfont (Figure~\ref{summary-explorer-examples}, Case A)}
Hallucinations are visualized by highlighting novel words in a summary. For each summary sentence with a hallucination, semantically and lexically similar document sentences are highlighted on demand. Since named entities and thus also entity relations form a subset of hallucinated words, the above coverage views do the same. Also, in an aggregated view, hallucinations found in multiple summaries are ordered by frequency, allowing to inspect a particular model with respect to types of hallucinations.

\paragraph{Position Bias View \normalfont (Figure~\ref{summary-explorer-views}e,f)}
Position bias is visualized for all models given a source document, and for a specific model with respect to all its summaries in a corpus. The former is visualized as a text heatmap, where a gradient color indicates for every sentence in a source document how many different summaries contain a semantically or lexically corresponding sentence. The latter is visualized by a different kind of heatmap for 50~randomly selected model summaries, where each summary is projected on a single horizontal bar representing the source document. Bar length reflects document length in sentences and aligned sentences are colored to reflect lexical or semantic alignment.

\paragraph{Aggregation Options}
Most of the above visualizations show individual pairs of source documents and a summary. This enables the close inspection of a given summary, and thus the manual assessment of a model by sequentially inspecting a number of summaries for different source documents generated by the same model. For these views, the visualizations also support displaying a number of summaries from different models for a relative assessment of their summaries.
\section{Collection of Model Outputs}

We collected the outputs of 55~summarization approaches on the test sets of three benchmark datasets for the task of single document summarization: CNN/DM, XSum and Webis-TLDR-17. Each dataset has a different style of ground truth summaries, ranging from semi-extractive to highly abstractive, providing a diverse selection of models. Outputs were obtained from NLPProgress, meta-evaluations such as SummEval \cite{fabbri:2021}, REALSumm \cite{bhandari:2020}, and in correspondence with the model's developers.%
\footnote{We sincerely thank all the developers for their efforts to reproduce and share their models' outputs with us.}

\subsection{Summarization Corpora}

The most popular dataset, CNN/DM \cite{hermann:2015,nallapati:2016}, contains news articles with multi-sentence summaries that are mostly extractive in nature \cite{kryscinski:2019,bommasani:2020}. We obtained the outputs from 45~models. While the original test split of the dataset contained 11,493~articles, we discarded ones that were not summarized by all models, resulting in 11,448~articles total. This minor discrepancy is due to inconsistent usage by authors, such as reshuffling the order of examples, de-duplication of articles in the test set, choice of tokenization, text capitalization, and truncation.

\enlargethispage{\baselineskip}
For the XSum dataset \cite{narayan:2018}, the outputs of six models for its test split (10,360~articles) were obtained. XSum contains news articles with more abstractive single-sentence summaries compared to CNN/DM.
The Webis-TLDR-17 dataset \cite{voelske:2017} contains highly abstractive, self-authored (single to multi-sentence) summaries of Reddit posts, although slightly noisier than the other datasets \cite{bommasani:2020}. We obtained the outputs from the four submissions of the TL;DR challenge \cite{syed:2019} for 250~posts.

\hwfigure[width=\textwidth]{summary-explorer-examples}{Two showcases for identifying inconsistencies in abstractive summaries using {\small\textsc{Summary Explorer}}. Case~A depicts the verification of the correctness of hallucinations by aligning document sentences. Case~B depicts uncovering more subtle hallucination errors by comparing unaligned relations.\vspace{-2ex}}

\subsection{Text Preprocessing}
\label{text-processing}

In a preprocessing pipeline, the input of a collection of documents, their ground truth summaries, and the generated summaries from a given model were normalized. First, basic normalization, such as de-tokenization, unifying model-specific sentence delimiters, and sentence segmentation were carried out. Second, additional information, such as named entities and relations were extracted using Spacy%
\footnote{\url{https://spacy.io}}
and Stanford OpenIE \cite{angeli:2015}, respectively. The latter extracts redundant relations where partial components such as either the subject or the object are already captured by longer counterparts. Such ``contained'' relations are merged into unique representative relations for each subject.

\enlargethispage{\baselineskip}
\paragraph{Alignment}
Every output summary is aligned with its source document, identifying the top two lexically and semantically related document sentences for each summary sentence. Lexical alignment relies on averaged ROUGE-$\{$1,2,L$\}$ scores among the document and summary sentences. The highest scoring document sentence is taken as the first match. The second match is identified by removing all content words from the summary sentence already captured by the first match, and repeating the process as per \citet{lebanoff:2019}. For semantic alignment, the rescaled BERTScore \cite{zhang:2020a} is computed between a summary sentence and all source document sentences, with the top-scoring two sentences as candidates.

\paragraph{Summary Evaluation Measures}
Several standard evaluation measures enable quantitative comparisons and filtering of models for detailed analysis:
\Ni
\emph{compression} as the word ratio between a document and its summary \cite{grusky:2018},
\Nii
\emph{n-gram abstractiveness} as per \citet{gehrmann:2019} calculates a normalized score for novelty by tracking parts of a summary that are already among the n-grams it has in common with its document,
\Niii
\emph{summary length} as word count (not tokens),
\Niv
\emph{entity-level factuality} as per \cite{nan:2021} as percentage of named entities in a summary found in its source document, and
\Nv
\emph{relation-level factuality} as percentage of relations in a summary found in its source document.
Finally, for consistency, we recompute ROUGE-$\{$1,2,L$\}$%
\footnote{\url{https://github.com/google-research/google-research/tree/master/rouge}}
for all the models.

\section{Assessment Case Studies}
\label{analysis}

We showcase the use and effectiveness of {\small\textsc{Summary Explorer}} by investigating two models ({\small\textsf{IMPROVE-ABS-NOVELTY}}, and {\small\textsf{IMPROVE-ABS-NOVELTY-LM}}) from \citet{kryscinski:2018} that improve the abstraction in summaries by including more novel phrases. We investigate the correctness of their hallucinations (novel words in the summary), and identify hidden errors introduced by the sentence fusion of the abstractive models.

\paragraph{Hallucinations via Sentence Alignment}
Hallucinations are novel words or phrases in a summary that warrant further inspection. Accordingly, our tool highlights them (Figure~\ref{summary-explorer-examples}, Case~A), directing the user to the respective candidate summary sentences whose related document sentences can be seen on demand. For {\small\textsf{IMPROVE-ABS-NOVELTY}}, we see that the first candidate improves abstraction via paraphrasing, is concisely written, and correctly substitutes the term \emph{``offenses''} with the novel word \emph{``charges''}. The second candidate also improves abstraction via sentence fusion, where two pieces of information are combined: \emph{``bennett allegedly drove her daughter''}, and \emph{``victim advised she thought she was going to die''}. The novel word \emph{``told''} also fits. However, the sentence fusion creates a wrong relation between the different actors (\emph{``bennett allegedly told her daughter that she was going to die''}), which can be easily identified via the visual sentence alignment provided.

\paragraph{Hidden Errors via Relation Alignment}
The above showcase does not capture all hallucinations. {\small\textsc{Summary Explorer}} also aligns relations extracted from a summary and its source document to identify novel relations. For {\small\textsf{IMPROVE-ABS-NOVELTY-LM}}, we see that the relation \emph{``she was arrested''} is unaligned to any relation in the source document (Figure \ref{summary-explorer-examples}, Case~B). Aligning the summary sentence to the document, we note that it is unfaithful to the source despite avoiding hallucinations (\emph{``Bennett was released on \$10,500~bail''}, and not \emph{``arrested on \$10,500~bail''}). The word \emph{``arrested''} was simply extracted from the document sentence (Figure~\ref{summary-explorer-examples}, Case~A). Without the visual support, identifying this small but important mistake would have been more cognitively demanding for an assessor.

\section{Conclusion}

In this paper, we present {\small\textsc{Summary Explorer}}, an online interactive visualization tool to assess the state of the art in text summarization in a guided fashion. In enables analysis akin to close and distant reading in particular facilitating the challenging inspection of hallucinations by abstractive summarization models. The tool is available open source%
\footnote{\url{https://github.com/webis-de/summary-explorer}}
enabling local use. We also welcome submissions of summaries from newer models trained on the existing datasets as part of our collaboration with the summarization community.
We aim to expand the tool's features in future work, exploring novel visual comparisons of documents to their summaries for more reliable qualitative assessments of summary quality. Finally, it is important to note that the accuracy of some of the views is influenced by the intrinsic drawbacks of the toolkits used for named entity recognition and information extraction.

\section{Ethical Statement}

Visualization plays a major role in the usage and accessibility of our tool. In this regard, to accommodate for color blindness, we primarily use gradient-based visuals for key modules such as model selection, aggregating important content, and text alignment. This renders the tool usable also in a monochromatic setting. Regarding the hosted summarization models, the key goal is to allow a wider audience comprising of model developers, the end users, and practitioners to openly compare and assess the strengths, limitations and possible ethical biases of these systems. Here, our tool supports making informed decisions about the suitability of certain models to the downstream applications. 

\section*{Acknowledgments}

We thank the reviewers for their valuable feedback. This work was supported by the German Federal Ministry of Education and Research (BMBF, 01/S18026A-F) by funding the competence center for Big Data and AI (ScaDS.AI Dresden/Leipzig).

\bibliography{emnlp21-tldr-demo}

\begin{thebibliography}{45}
\expandafter\ifx\csname natexlab\endcsname\relax\def\natexlab#1{#1}\fi

\bibitem[{Angeli et~al.(2015)Angeli, Premkumar, and Manning}]{angeli:2015}
Gabor Angeli, Melvin Jose~Johnson Premkumar, and Christopher~D. Manning. 2015.
\newblock \href {https://doi.org/10.3115/v1/p15-1034} {Leveraging linguistic
  structure for open domain information extraction}.
\newblock In \emph{Proceedings of the 53rd Annual Meeting of the Association
  for Computational Linguistics and the 7th International Joint Conference on
  Natural Language Processing of the Asian Federation of Natural Language
  Processing, {ACL} 2015, July 26-31, 2015, Beijing, China, Volume 1: Long
  Papers}, pages 344--354. The Association for Computer Linguistics.

\bibitem[{Bhandari et~al.(2020)Bhandari, Gour, Ashfaq, Liu, and
  Neubig}]{bhandari:2020}
Manik Bhandari, Pranav~Narayan Gour, Atabak Ashfaq, Pengfei Liu, and Graham
  Neubig. 2020.
\newblock \href {https://www.aclweb.org/anthology/2020.emnlp-main.751}
  {Re-evaluating evaluation in text summarization}.
\newblock In \emph{Proceedings of the 2020 Conference on Empirical Methods in
  Natural Language Processing (EMNLP)}, Online. Association for Computational
  Linguistics.

\bibitem[{Bommasani and Cardie(2020)}]{bommasani:2020}
Rishi Bommasani and Claire Cardie. 2020.
\newblock \href {https://doi.org/10.18653/v1/2020.emnlp-main.649} {Intrinsic
  evaluation of summarization datasets}.
\newblock In \emph{Proceedings of the 2020 Conference on Empirical Methods in
  Natural Language Processing (EMNLP)}, pages 8075--8096, Online. Association
  for Computational Linguistics.

\bibitem[{Cao et~al.(2018)Cao, Wei, Li, and Li}]{cao:2018}
Ziqiang Cao, Furu Wei, Wenjie Li, and Sujian Li. 2018.
\newblock \href
  {https://www.aaai.org/ocs/index.php/AAAI/AAAI18/paper/view/16121} {Faithful
  to the original: Fact aware neural abstractive summarization}.
\newblock In \emph{Proceedings of the Thirty-Second {AAAI} Conference on
  Artificial Intelligence, (AAAI-18), the 30th innovative Applications of
  Artificial Intelligence (IAAI-18), and the 8th {AAAI} Symposium on
  Educational Advances in Artificial Intelligence (EAAI-18), New Orleans,
  Louisiana, USA, February 2-7, 2018}, pages 4784--4791. {AAAI} Press.

\bibitem[{Dong et~al.(2018)Dong, Shen, Crawford, van Hoof, and
  Cheung}]{dong:2018}
Yue Dong, Yikang Shen, Eric Crawford, Herke van Hoof, and Jackie Chi~Kit
  Cheung. 2018.
\newblock \href {https://doi.org/10.18653/v1/d18-1409} {Banditsum: Extractive
  summarization as a contextual bandit}.
\newblock In \emph{Proceedings of the 2018 Conference on Empirical Methods in
  Natural Language Processing, Brussels, Belgium, October 31 - November 4,
  2018}, pages 3739--3748. Association for Computational Linguistics.

\bibitem[{Durmus et~al.(2020)Durmus, He, and Diab}]{durmus:2020}
Esin Durmus, He~He, and Mona Diab. 2020.
\newblock \href {https://doi.org/10.18653/v1/2020.acl-main.454} {{FEQA:} {A}
  question answering evaluation framework for faithfulness assessment in
  abstractive summarization}.
\newblock In \emph{Proceedings of the 58th Annual Meeting of the Association
  for Computational Linguistics, {ACL} 2020, Online, July 5-10, 2020}, pages
  5055--5070. Association for Computational Linguistics.

\bibitem[{Fabbri et~al.(2021)Fabbri, Kryscinski, McCann, Xiong, Socher, and
  Radev}]{fabbri:2021}
Alexander~R. Fabbri, Wojciech Kryscinski, Bryan McCann, Caiming Xiong, Richard
  Socher, and Dragomir~R. Radev. 2021.
\newblock \href {https://transacl.org/ojs/index.php/tacl/article/view/2563}
  {Summeval: Re-evaluating summarization evaluation}.
\newblock \emph{Trans. Assoc. Comput. Linguistics}, 9:391--409.

\bibitem[{Falke et~al.(2019)Falke, Ribeiro, Utama, Dagan, and
  Gurevych}]{falke:2019}
Tobias Falke, Leonardo F.~R. Ribeiro, Prasetya~Ajie Utama, Ido Dagan, and Iryna
  Gurevych. 2019.
\newblock \href {https://doi.org/10.18653/v1/p19-1213} {Ranking generated
  summaries by correctness: An interesting but challenging application for
  natural language inference}.
\newblock In \emph{Proceedings of the 57th Conference of the Association for
  Computational Linguistics, {ACL} 2019, Florence, Italy, July 28- August 2,
  2019, Volume 1: Long Papers}, pages 2214--2220. Association for Computational
  Linguistics.

\bibitem[{Gehrmann et~al.(2018)Gehrmann, Deng, and Rush}]{gehrmann:2018}
Sebastian Gehrmann, Yuntian Deng, and Alexander~M. Rush. 2018.
\newblock \href {https://doi.org/10.18653/v1/d18-1443} {Bottom-up abstractive
  summarization}.
\newblock In \emph{Proceedings of the 2018 Conference on Empirical Methods in
  Natural Language Processing, Brussels, Belgium, October 31 - November 4,
  2018}, pages 4098--4109. Association for Computational Linguistics.

\bibitem[{Gehrmann et~al.(2019)Gehrmann, Ziegler, and Rush}]{gehrmann:2019}
Sebastian Gehrmann, Zachary~M. Ziegler, and Alexander~M. Rush. 2019.
\newblock \href {https://aclweb.org/anthology/papers/W/W19/W19-8665/}
  {Generating abstractive summaries with finetuned language models}.
\newblock In \emph{Proceedings of the 12th International Conference on Natural
  Language Generation, {INLG} 2019, Tokyo, Japan, October 29 - November 1,
  2019}, pages 516--522. Association for Computational Linguistics.

\bibitem[{Grusky et~al.(2018)Grusky, Naaman, and Artzi}]{grusky:2018}
Max Grusky, Mor Naaman, and Yoav Artzi. 2018.
\newblock \href {https://doi.org/10.18653/v1/n18-1065} {Newsroom: {A} dataset
  of 1.3 million summaries with diverse extractive strategies}.
\newblock In \emph{Proceedings of the 2018 Conference of the North American
  Chapter of the Association for Computational Linguistics: Human Language
  Technologies, {NAACL-HLT} 2018, New Orleans, Louisiana, USA, June 1-6, 2018,
  Volume 1 (Long Papers)}, pages 708--719. Association for Computational
  Linguistics.

\bibitem[{Hardy et~al.(2019)Hardy, Narayan, and Vlachos}]{hardy:2019}
Hardy, Shashi Narayan, and Andreas Vlachos. 2019.
\newblock \href {https://doi.org/10.18653/v1/p19-1330} {Highres:
  Highlight-based reference-less evaluation of summarization}.
\newblock In \emph{Proceedings of the 57th Conference of the Association for
  Computational Linguistics, {ACL} 2019, Florence, Italy, July 28- August 2,
  2019, Volume 1: Long Papers}, pages 3381--3392. Association for Computational
  Linguistics.

\bibitem[{Hermann et~al.(2015)Hermann, Kocisk{\'{y}}, Grefenstette, Espeholt,
  Kay, Suleyman, and Blunsom}]{hermann:2015}
Karl~Moritz Hermann, Tom{\'{a}}s Kocisk{\'{y}}, Edward Grefenstette, Lasse
  Espeholt, Will Kay, Mustafa Suleyman, and Phil Blunsom. 2015.
\newblock \href
  {http://papers.nips.cc/paper/5945-teaching-machines-to-read-and-comprehend}
  {Teaching machines to read and comprehend}.
\newblock In \emph{Advances in Neural Information Processing Systems 28: Annual
  Conference on Neural Information Processing Systems 2015, December 7-12,
  2015, Montreal, Quebec, Canada}, pages 1693--1701.

\bibitem[{Jones et~al.(1999)}]{jones:1999}
K~Sparck Jones et~al. 1999.
\newblock Automatic summarizing: factors and directions.
\newblock \emph{Advances in automatic text summarization}, pages 1--12.

\bibitem[{Kedzie et~al.(2018)Kedzie, McKeown, and III}]{kedzie:2018}
Chris Kedzie, Kathleen~R. McKeown, and Hal~Daum{\'{e}} III. 2018.
\newblock \href {https://doi.org/10.18653/v1/d18-1208} {Content selection in
  deep learning models of summarization}.
\newblock In \emph{Proceedings of the 2018 Conference on Empirical Methods in
  Natural Language Processing, Brussels, Belgium, October 31 - November 4,
  2018}, pages 1818--1828. Association for Computational Linguistics.

\bibitem[{Kryscinski et~al.(2019)Kryscinski, Keskar, McCann, Xiong, and
  Socher}]{kryscinski:2019}
Wojciech Kryscinski, Nitish~Shirish Keskar, Bryan McCann, Caiming Xiong, and
  Richard Socher. 2019.
\newblock \href {https://doi.org/10.18653/v1/D19-1051} {Neural text
  summarization: {A} critical evaluation}.
\newblock In \emph{Proceedings of the 2019 Conference on Empirical Methods in
  Natural Language Processing and the 9th International Joint Conference on
  Natural Language Processing, {EMNLP-IJCNLP} 2019, Hong Kong, China, November
  3-7, 2019}, pages 540--551. Association for Computational Linguistics.

\bibitem[{Kryscinski et~al.(2020)Kryscinski, McCann, Xiong, and
  Socher}]{kryscinski:2020}
Wojciech Kryscinski, Bryan McCann, Caiming Xiong, and Richard Socher. 2020.
\newblock \href {https://doi.org/10.18653/v1/2020.emnlp-main.750} {Evaluating
  the factual consistency of abstractive text summarization}.
\newblock In \emph{Proceedings of the 2020 Conference on Empirical Methods in
  Natural Language Processing, {EMNLP} 2020, Online, November 16-20, 2020},
  pages 9332--9346. Association for Computational Linguistics.

\bibitem[{Kryscinski et~al.(2018)Kryscinski, Paulus, Xiong, and
  Socher}]{kryscinski:2018}
Wojciech Kryscinski, Romain Paulus, Caiming Xiong, and Richard Socher. 2018.
\newblock \href {https://doi.org/10.18653/v1/d18-1207} {Improving abstraction
  in text summarization}.
\newblock In \emph{Proceedings of the 2018 Conference on Empirical Methods in
  Natural Language Processing, Brussels, Belgium, October 31 - November 4,
  2018}, pages 1808--1817. Association for Computational Linguistics.

\bibitem[{Lebanoff et~al.(2019)Lebanoff, Song, Dernoncourt, Kim, Kim, Chang,
  and Liu}]{lebanoff:2019}
Logan Lebanoff, Kaiqiang Song, Franck Dernoncourt, Doo~Soon Kim, Seokhwan Kim,
  Walter Chang, and Fei Liu. 2019.
\newblock \href {https://doi.org/10.18653/v1/p19-1209} {Scoring sentence
  singletons and pairs for abstractive summarization}.
\newblock In \emph{Proceedings of the 57th Conference of the Association for
  Computational Linguistics, {ACL} 2019, Florence, Italy, July 28- August 2,
  2019, Volume 1: Long Papers}, pages 2175--2189. Association for Computational
  Linguistics.

\bibitem[{Lin(2004)}]{lin:2004}
Chin-Yew Lin. 2004.
\newblock \href {https://www.aclweb.org/anthology/W04-1013} {{ROUGE}: A package
  for automatic evaluation of summaries}.
\newblock In \emph{Text Summarization Branches Out}, pages 74--81, Barcelona,
  Spain. Association for Computational Linguistics.

\bibitem[{Lloret et~al.(2018)Lloret, Plaza, and Aker}]{lloret:2018}
Elena Lloret, Laura Plaza, and Ahmet Aker. 2018.
\newblock \href {https://doi.org/10.1007/s10579-017-9399-2} {The challenging
  task of summary evaluation: an overview}.
\newblock \emph{Lang. Resour. Evaluation}, 52(1):101--148.

\bibitem[{Louis and Nenkova(2013)}]{nenkova:2013}
Annie Louis and Ani Nenkova. 2013.
\newblock \href {https://doi.org/10.1162/COLI\_a\_00123} {Automatically
  assessing machine summary content without a gold standard}.
\newblock \emph{Comput. Linguistics}, 39(2):267--300.

\bibitem[{Luhn(1958)}]{luhn:1958}
Hans~Peter Luhn. 1958.
\newblock The automatic creation of literature abstracts.
\newblock \emph{{IBM} Journal of Reseach and Devopment}, 2(2):159--165.

\bibitem[{Mani(2001)}]{mani:2001}
Inderjeet Mani. 2001.
\newblock Summarization evaluation: An overview.

\bibitem[{Mani et~al.(1999)Mani, House, Klein, Hirschman, Firmin, and
  Sundheim}]{mani:1999}
Inderjeet Mani, David House, Gary Klein, Lynette Hirschman, Therese Firmin, and
  Beth Sundheim. 1999.
\newblock \href {https://www.aclweb.org/anthology/E99-1011/} {The tipster
  summac text summarization evaluation}.
\newblock In \emph{{EACL} 1999, 9th Conference of the European Chapter of the
  Association for Computational Linguistics, June 8-12, 1999, University of
  Bergen, Bergen, Norway}, pages 77--85. The Association for Computer
  Linguistics.

\bibitem[{Maynez et~al.(2020)Maynez, Narayan, Bohnet, and
  McDonald}]{maynez:2020}
Joshua Maynez, Shashi Narayan, Bernd Bohnet, and Ryan~T. McDonald. 2020.
\newblock \href {https://doi.org/10.18653/v1/2020.acl-main.173} {On
  faithfulness and factuality in abstractive summarization}.
\newblock In \emph{Proceedings of the 58th Annual Meeting of the Association
  for Computational Linguistics, {ACL} 2020, Online, July 5-10, 2020}, pages
  1906--1919. Association for Computational Linguistics.

\bibitem[{Nallapati et~al.(2016)Nallapati, Zhou, dos Santos,
  G{\"{u}}l{\c{c}}ehre, and Xiang}]{nallapati:2016}
Ramesh Nallapati, Bowen Zhou, C{\'{\i}}cero~Nogueira dos Santos, {\c{C}}aglar
  G{\"{u}}l{\c{c}}ehre, and Bing Xiang. 2016.
\newblock \href {https://doi.org/10.18653/v1/k16-1028} {Abstractive text
  summarization using sequence-to-sequence rnns and beyond}.
\newblock In \emph{Proceedings of the 20th {SIGNLL} Conference on Computational
  Natural Language Learning, CoNLL 2016, Berlin, Germany, August 11-12, 2016},
  pages 280--290. {ACL}.

\bibitem[{Nan et~al.(2021)Nan, Nallapati, Wang, dos Santos, Zhu, Zhang,
  McKeown, and Xiang}]{nan:2021}
Feng Nan, Ramesh Nallapati, Zhiguo Wang, C{\'{\i}}cero~Nogueira dos Santos,
  Henghui Zhu, Dejiao Zhang, Kathy McKeown, and Bing Xiang. 2021.
\newblock \href {https://www.aclweb.org/anthology/2021.eacl-main.235/}
  {Entity-level factual consistency of abstractive text summarization}.
\newblock In \emph{Proceedings of the 16th Conference of the European Chapter
  of the Association for Computational Linguistics: Main Volume, {EACL} 2021,
  Online, April 19 - 23, 2021}, pages 2727--2733. Association for Computational
  Linguistics.

\bibitem[{Narayan et~al.(2018)Narayan, Cohen, and Lapata}]{narayan:2018}
Shashi Narayan, Shay~B. Cohen, and Mirella Lapata. 2018.
\newblock \href {https://doi.org/10.18653/v1/d18-1206} {Don't give me the
  details, just the summary! topic-aware convolutional neural networks for
  extreme summarization}.
\newblock In \emph{Proceedings of the 2018 Conference on Empirical Methods in
  Natural Language Processing, Brussels, Belgium, October 31 - November 4,
  2018}, pages 1797--1807. Association for Computational Linguistics.

\bibitem[{Ng and Abrecht(2015)}]{ng:2015}
Jun{-}Ping Ng and Viktoria Abrecht. 2015.
\newblock \href {https://doi.org/10.18653/v1/d15-1222} {Better summarization
  evaluation with word embeddings for {ROUGE}}.
\newblock In \emph{Proceedings of the 2015 Conference on Empirical Methods in
  Natural Language Processing, {EMNLP} 2015, Lisbon, Portugal, September 17-21,
  2015}, pages 1925--1930. The Association for Computational Linguistics.

\bibitem[{Paice(1990)}]{paice:1990}
Chris~D Paice. 1990.
\newblock Constructing literature abstracts by computer: techniques and
  prospects.
\newblock \emph{Information Processing \& Management}, 26(1):171--186.

\bibitem[{Peyrard(2019)}]{peyrard:2019}
Maxime Peyrard. 2019.
\newblock \href {https://doi.org/10.18653/v1/p19-1101} {A simple theoretical
  model of importance for summarization}.
\newblock In \emph{Proceedings of the 57th Conference of the Association for
  Computational Linguistics, {ACL} 2019, Florence, Italy, July 28- August 2,
  2019, Volume 1: Long Papers}, pages 1059--1073. Association for Computational
  Linguistics.

\bibitem[{PurdueOWL(2019)}]{purdue:2019}
PurdueOWL. 2019.
\newblock \href
  {http://web.archive.org/web/20080207010024/http://www.808multimedia.com/winnt/kernel.htm}
  {Journalism and journalistic writing: The inverted pyramid structure}.

\bibitem[{Salton et~al.(1994)Salton, Allan, Buckley, and Singhal}]{salton:1994}
Gerard Salton, James Allan, Chris Buckley, and Amit Singhal. 1994.
\newblock Automatic analysis, theme generation, and summarization of
  machine-readable texts.
\newblock \emph{Science}, 264(5164):1421--1426.

\bibitem[{ShafieiBavani et~al.(2018)ShafieiBavani, Ebrahimi, Wong, and
  Chen}]{shafieibavani:2018}
Elaheh ShafieiBavani, Mohammad Ebrahimi, Raymond~K. Wong, and Fang Chen. 2018.
\newblock \href {https://www.aclweb.org/anthology/C18-1077/} {Summarization
  evaluation in the absence of human model summaries using the compositionality
  of word embeddings}.
\newblock In \emph{Proceedings of the 27th International Conference on
  Computational Linguistics, {COLING} 2018, Santa Fe, New Mexico, USA, August
  20-26, 2018}, pages 905--914. Association for Computational Linguistics.

\bibitem[{Shneiderman(1996)}]{shneiderman:1996}
B.~Shneiderman. 1996.
\newblock \href {https://doi.org/10.1109/VL.1996.545307} {The eyes have it: a
  task by data type taxonomy for information visualizations}.
\newblock In \emph{Proceedings 1996 IEEE Symposium on Visual Languages}, pages
  336--343.

\bibitem[{Syed et~al.(2019)Syed, V{\"{o}}lske, Lipka, Stein, Sch{\"{u}}tze, and
  Potthast}]{syed:2019}
Shahbaz Syed, Michael V{\"{o}}lske, Nedim Lipka, Benno Stein, Hinrich
  Sch{\"{u}}tze, and Martin Potthast. 2019.
\newblock \href {https://aclweb.org/anthology/papers/W/W19/W19-8666/} {Towards
  summarization for social media - results of the tl;dr challenge}.
\newblock In \emph{Proceedings of the 12th International Conference on Natural
  Language Generation, {INLG} 2019, Tokyo, Japan, October 29 - November 1,
  2019}, pages 523--528. Association for Computational Linguistics.

\bibitem[{Tenney et~al.(2020)Tenney, Wexler, Bastings, Bolukbasi, Coenen,
  Gehrmann, Jiang, Pushkarna, Radebaugh, Reif, and Yuan}]{tenney:2020}
Ian Tenney, James Wexler, Jasmijn Bastings, Tolga Bolukbasi, Andy Coenen,
  Sebastian Gehrmann, Ellen Jiang, Mahima Pushkarna, Carey Radebaugh, Emily
  Reif, and Ann Yuan. 2020.
\newblock \href {https://doi.org/10.18653/v1/2020.emnlp-demos.15} {The language
  interpretability tool: Extensible, interactive visualizations and analysis
  for {NLP} models}.
\newblock In \emph{Proceedings of the 2020 Conference on Empirical Methods in
  Natural Language Processing: System Demonstrations, {EMNLP} 2020 - Demos,
  Online, November 16-20, 2020}, pages 107--118. Association for Computational
  Linguistics.

\bibitem[{Vig et~al.(2021)Vig, Kryscinski, Goel, and Rajani}]{vig:2021}
Jesse Vig, Wojciech Kryscinski, Karan Goel, and Nazneen~Fatema Rajani. 2021.
\newblock \href {http://arxiv.org/abs/2104.07605} {Summvis: Interactive visual
  analysis of models, data, and evaluation for text summarization}.
\newblock \emph{CoRR}, abs/2104.07605.

\bibitem[{V{\"{o}}lske et~al.(2017)V{\"{o}}lske, Potthast, Syed, and
  Stein}]{voelske:2017}
Michael V{\"{o}}lske, Martin Potthast, Shahbaz Syed, and Benno Stein. 2017.
\newblock \href {https://doi.org/10.18653/v1/w17-4508} {Tl;dr: Mining reddit to
  learn automatic summarization}.
\newblock In \emph{Proceedings of the Workshop on New Frontiers in
  Summarization, NFiS@EMNLP 2017, Copenhagen, Denmark, September 7, 2017},
  pages 59--63. Association for Computational Linguistics.

\bibitem[{Wang et~al.(2019)Wang, Jain, Chen, and Gu}]{wang:2019}
Changhan Wang, Anirudh Jain, Danlu Chen, and Jiatao Gu. 2019.
\newblock \href {https://doi.org/10.18653/v1/D19-3043} {Vizseq: a visual
  analysis toolkit for text generation tasks}.
\newblock In \emph{Proceedings of the 2019 Conference on Empirical Methods in
  Natural Language Processing and the 9th International Joint Conference on
  Natural Language Processing, {EMNLP-IJCNLP} 2019, Hong Kong, China, November
  3-7, 2019 - System Demonstrations}, pages 253--258. Association for
  Computational Linguistics.

\bibitem[{Wasson(1998)}]{wasson:1998}
Mark Wasson. 1998.
\newblock Using leading text for news summaries: Evaluation results and
  implications for commercial summarization applications.
\newblock In \emph{Proc. of the 36th Annual Meeting of the Association for
  Computational Linguistics and 17th Int. Conference on Computational
  Linguistics}, pages 1364--1368.

\bibitem[{Yousef and J{\"a}nicke(2021)}]{yousef:2021}
Tariq Yousef and Stefan J{\"a}nicke. 2021.
\newblock \href {https://doi.org/10.1109/TVCG.2020.3028975} {A survey of text
  alignment visualization}.
\newblock \emph{IEEE Transactions on Visualization and Computer Graphics},
  27(2):1149--1159.

\bibitem[{Zhang et~al.(2020)Zhang, Kishore, Wu, Weinberger, and
  Artzi}]{zhang:2020a}
Tianyi Zhang, Varsha Kishore, Felix Wu, Kilian~Q. Weinberger, and Yoav Artzi.
  2020.
\newblock \href {https://openreview.net/forum?id=SkeHuCVFDr} {Bertscore:
  Evaluating text generation with {BERT}}.
\newblock In \emph{8th International Conference on Learning Representations,
  {ICLR} 2020, Addis Ababa, Ethiopia, April 26-30, 2020}. OpenReview.net.

\bibitem[{Zhao et~al.(2020)Zhao, Cohen, and Webber}]{zhao:2020}
Zheng Zhao, Shay~B. Cohen, and Bonnie Webber. 2020.
\newblock \href {https://www.aclweb.org/anthology/2020.findings-emnlp.203}
  {Reducing quantity hallucinations in abstractive summarization}.
\newblock In \emph{Findings of the Association for Computational Linguistics:
  EMNLP 2020}, Online. Association for Computational Linguistics.

\end{thebibliography}
\bibliographystyle{acl_natbib}
\end{document}